# THE DATA SCIENCE OF HOLLYWOOD: USING EMOTIONAL ARCS OF MOVIES TO DRIVE BUSINESS MODEL INNOVATION IN ENTERTAINMENT INDUSTRIES*

Marco Del Vecchio†      Alexander Kharlamov#

Glenn Parry‡      Ganna Pogrebna§

June 2018

## Abstract

Much of business literature addresses the issues of consumer-centric design: how can businesses design customized services and products which accurately reflect consumer preferences? This paper uses data science natural language processing methodology to explore whether and to what extent emotions shape consumer preferences for media and entertainment content. Using a unique filtered dataset of 6,174 movie scripts, we generate a mapping of screen content to capture the emotional trajectory of each motion picture. We then combine the obtained mappings into clusters which represent groupings of consumer emotional journeys. These clusters are used to predict overall success parameters of the movies including box office revenues, viewer satisfaction levels (captured by IMDb ratings), awards, as well as the number of viewers' and critics' reviews. We find that like books all movie stories are dominated by 6 basic shapes. The highest box offices are associated with the *Man in a Hole* shape which is characterized by an emotional fall followed by an emotional rise. This shape results in financially successful movies irrespective of genre and production budget. Yet, *Man in a Hole* succeeds not because it produces most "liked" movies but because it generates most "talked about" movies. Interestingly, a carefully chosen combination of production budget and genre may produce a financially successful movie with any emotional shape. Implications of this analysis for generating on-demand content and for driving business model innovation in entertainment industries are discussed.

**Keywords:** sentiment analysis, sentiment mining, consumer-centric design, entertainment, emotions, storytelling

* The first version of this paper was presented at the research seminar at Warwick Manufacturing Group in July 2017 under the title "Sentiment-driven Consumer-centric Design: Understanding Emotional Trajectories of Films to Drive Business Model Innovation in Media and Entertainment Industries". We are grateful to the participants at research seminars at the University College London, University of York, University of East Anglia, Impact Showcase at the Birmingham Business School, the Data Science in Entertainment Industries Conference in Manchester in March 2018, and the Google Catalyst Summit in Dublin in May 2018 for many useful comments and suggestions. Ganna Pogrebna acknowledges financial support from RCUK/EPSRC grants EP/N028422/1 and EP/P011896/1. We thank Alexander Milanovic for excellent research assistance.

† University of Cambridge, Department of Engineering, Trumpington Street, Cambridge, CB2 1PZ
# University of the West of England, Frenchay Campus, Coldharbour Lane, Bristol BS16 1QY, UK
‡ University of the West of England, Frenchay Campus, Coldharbour Lane, Bristol BS16 1QY, UK
§ Corresponding author: The Alan Turing Institute, 96 Euston Rd, Kings Cross, London NW1 2DB and Department of Economics, Birmingham Business School, University of Birmingham, JG Smith Building, Birmingham, B15 2TT and, e-mail: gpogrebna@turing.ac.uk

# THE DATA SCIENCE OF HOLLYWOOD: USING EMOTIONAL ARCS OF MOTION PICTURES TO DRIVE BUSINESS MODEL INNOVATION IN ENTERTAINMENT INDUSTRIES

## 1 Introduction

Many people regard motion pictures to be an inherent part of their lifetime cultural journey. Regardless of what one calls it – a "film", a "movie", or a "picture" – people often have favorites which they remember from childhood, quote on a regular basis, or even use to mimic the style of the main characters. But why do some movies become an almost immediate success going viral around the globe while others are quickly forgotten? The motion picture production and distribution industry is not only a multi-billion dollar market generating over $120 billion annually[1]; it is also a great storytelling enterprise. The stories told by the motion pictures help people connect with the characters, relive their own experiences, and even escape their daily lives. In this paper, we explore whether and to what extent the success of stories told by motion pictures is defined by the emotional journey which these stories offer to the viewers and how understanding these emotional journeys can drive business models in the entertainment industry.

Since Aristotle, writers have grappled with the magic formula for storytelling success, trying to anticipate and design the most engaging stories (Aristotle, 1902). In "The Poetics of Aristotle", Aristotle proposed that sparking an emotional response is very important for telling a successful story as well as identified several story types for ancient poetry. Specifically, he argued: "*A perfect tragedy should, as we have seen, be arranged not on the simple but on the*

---

[1] According to statista.com the market size of the global movie production and distribution industry in 2017 was $124 billion. For more details, see https://www.statista.com/statistics/326011/movie-production-distribution-industry/

*complex plan. It should, moreover, imitate actions which excite pity and fear, this being the distinctive mark of tragic imitation*" (Aristotle, 1902, p. 45).

While for many centuries, the emotional content of stories was largely a subject of linguistic analysis in humanities' research, recent advances in Natural Language Processing (NLP) and computational narratology allow scientists to significantly advance the sentiment analysis of storytelling. One of the first examples of using information technology to analyze emotional content of stories belongs to Kurt Vonnegut. He not only coined the term "emotional arc" of a story, but also visualized it in a two-dimensional space defining it as a correspondence between the timing of the story ("Beginning-End") displayed on a horizontal axes, and its emotional journey ("Ill Fortune-Great Fortune") shown on a vertical axes (Vonnegut, 1981). More recently, the methodology of Aristotle and Vonnegut was extended and popularized by a team of researchers from the Computational Story Laboratory at the University of Vermont who used the NLP methodology to map emotional journeys of a filtered dataset consisting of 1,327 novels from Project Gutenberg's digital fiction collection and identified 6 emotional arcs which describe all those stories (Reagan et al., 2016). Reagan el al. (2016) showed that all analyzed novels could be partitioned into 6 clusters where each cluster represents a specific emotional trajectory (see Figure 1):

- *Rags to riches* – an emotional trajectory showing an ongoing emotional rise.
- *Riches to rags* – an emotional trajectory showing an ongoing emotional fall.
- *Man in a hole* – an emotional trajectory showing a fall followed by a rise.
- *Icarus* – an emotional trajectory showing a rise followed by a fall.
- *Cinderella* – an emotional trajectory showing a rise-fall-rise pattern

- *Oedipus* – an emotional trajectory showing a fall-rise-fall pattern

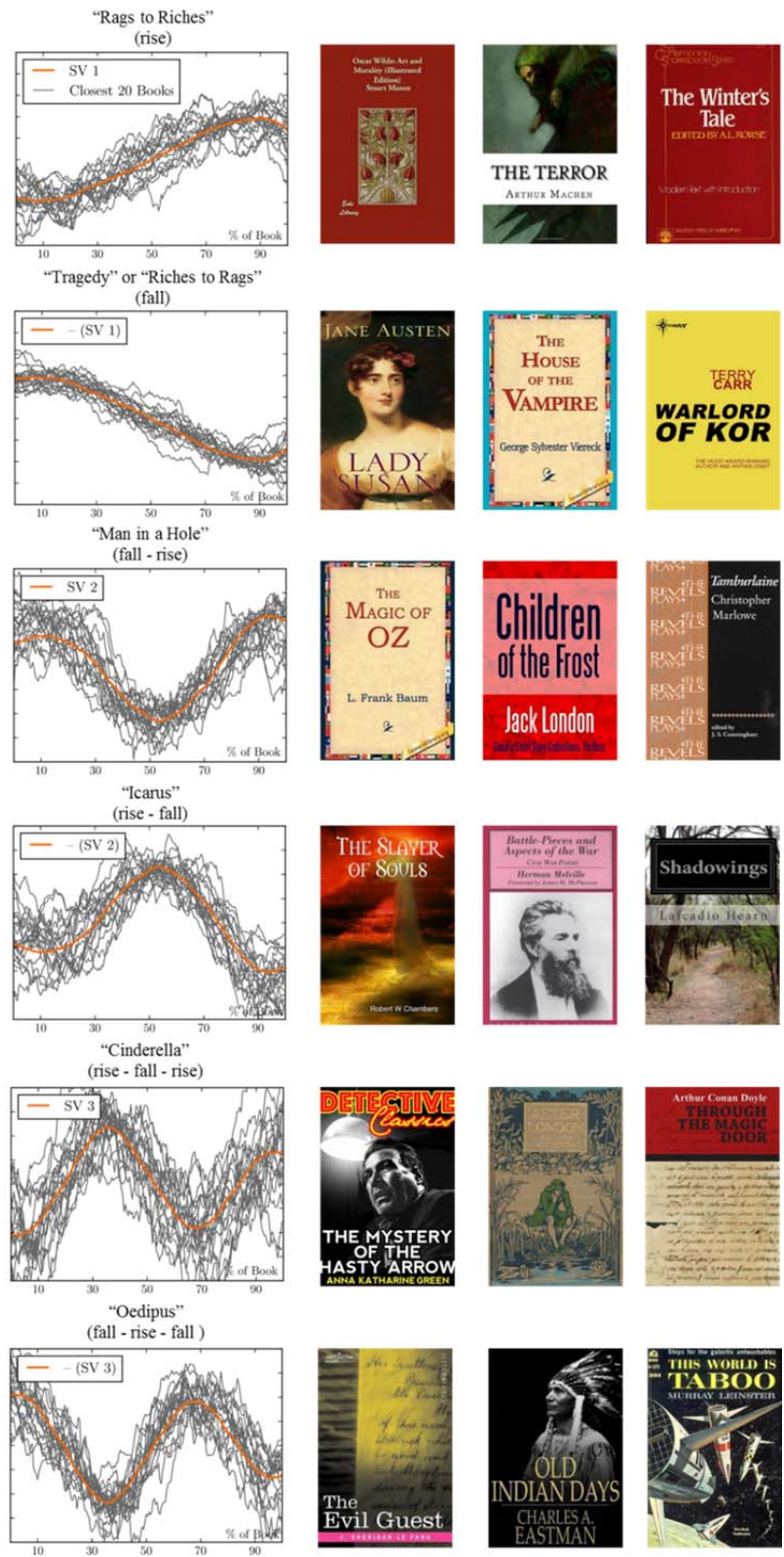

**Figure 1 Six Major Clusters of Emotional Trajectories for Novels with Examples Compiled from Reagan et al. (2016)**

Recently, the importance of emotional arcs has also been emphasized not only for storytelling (Fernandes et al., 2018; Ferraz de Arruda et al., 2018; Green et al., 2018; Grubert, and Algee-Hewitt, 2017) but also for the audio-visual content design (Chu and Roy, 2017). As award-winning scriptwriter Frank Cottrell-Boyce once put it talking about a recipe for a perfect motion picture story: "*All the manuals insist on a three-act structure. I think this is a useless model. It's static. All it really means is that your screenplay should have a beginning, middle and end. When you're shaping things, it's more useful to think about suspense. Suspense is the hidden energy that holds a story together. It connects two points and sends a charge between them. But it doesn't have to be all action. Emotions create their own suspense.*"[2] Using 509 Hollywood (full-length) motion pictures and 1,326 short videos from Vimeo channel "Short of the Week" (between 30 seconds and 30 minutes long), Chu and Roy (2017) combined audio and visual information from movies to map sentiment using neural networks methodology. For short videos, they identified audio-visual emotional arcs which attracted the highest number of comments on Vimeo. They showed that for videos with a median length of slightly over 8 minutes, the highest number of clicks were achieved by the emotional trajectory somewhat resembling *Icarus* which ended on a steep decline. Other trajectories with high number of clicks were characterized by significant emotional peaks close to the end of the video.

In this paper we use a unique filtered dataset of 6,174 full-length movie scripts from https://www.opensubtitles.org to generate a mapping of screen content capturing the emotional arc of each motion picture. We then accumulate emotional arcs into clustered trajectories which represent groupings of viewer emotional journeys. These clusters are then used to predict a wide variety of movie success characteristics: revenues, satisfaction levels, audience capture, award nominations and award wins.

---

[2] See *The Guardian* interview with Frank Cottrell-Boyce
https://www.theguardian.com/film/2008/jun/30/news.culture1 for more information.

We find that full-length motion pictures' scripts fall within the same 6 major emotional arcs as novels' arcs reported in Reagan et al. (2016). We also show that when success of a motion picture is measured by box office revenues, viewers tend to prefer movies with emotional trajectory of *Man in a Hole*. This result is robust even if we control for production budget and genre of the movie. We also conclude that *Man in a Hole* movies tend to succeed not because these motion pictures are associated with the highest viewer satisfaction. This emotional arc tends to attract viewers' attention and spark discussions. It does not mean, however, that only *Man in a Hole* movies are set for financial success: our results also show that if a genre and budget of the film is chosen carefully, it is possible to produce a financially successful movie in any of the 6 emotional arcs' shapes.

This paper is organized as follows. We start by describing our dataset in Section 2. Section 3 explains our research methodology, hypotheses, and analysis procedures. Results are described in Section 4. Finally, Section 5 concludes with a general discussion.

**2 The Data**

The dataset for this project was compiled from several sources. We harvested subtitle files from https://www.opensubtitles.org. Additional information about each motion picture was obtained from https://www.imdb.com. We also used https://www.the-numbers.com data on movies revenues as well as estimated production budgets.

In the first instance, 156,568 subtitle files were obtained from an open source website https://www.opensubtitles.org. As of June 25th 2018, the website had a collection of 4,524,139 subtitles in multiple languages.[3] For the purposes of this project we concentrated on subtitles in English. In order to filter the obtained subtitles for quality and reliability and make sure that the subtitle files were linked to our main proxy of success (revenue), we have applied the following procedure. First, if a motion picture had more than one subtitle file listed on https://www.opensubtitles.org, we removed duplicates and only kept files with the highest number of download count. This reduced the total number of subtitles to 27,883. Second, the obtained dataset was matched with the data extracted from https://www.the-numbers.com on revenues. This dataset was cross-checked and complimented with the data on revenues listed on https://www.imdb.com. The web resource https://www.the-numbers.com provided three variables for motion pictures[4]: estimated production budget, domestic gross revenue, and worldwide gross revenue. For the overwhelming majority of movies, domestic gross revenue meant domestic gross revenue in the US and measured in US dollars since the majority of movies in our sample were produced in the US. Where domestic gross revenue was indicated in British pounds or some other currency, we have converted the revenue number to US dollars. Domestic gross revenue was available for 9,015 motion pictures. Production budget estimates and worldwide gross revenues were available for a subset of these movies. We removed movies records for which we could not find domestic gross revenue, yielding 9,015 records.

Third, quality control criteria were applied to the dataset. The subtitles repository https://www.opensubtitles.org is an open-source website where individual users post subtitle files. Yet, it allows all subtitle consumers to rank user members who post subtitles awarding them bronze, silver, gold, or platinum membership ranks (see Figure 2). The membership rank depends on the quality of subtitles users post as downloaded and rated by other users. We only used subtitles from ranked users (bronze, silver, gold, and platinum members) and discarded scripts posted by unranked users reducing the dataset to 6,562 subtitle records. We then removed all subtitles where the length of the text was less than 10,000 characters to ensure that our analysis is based on long motion pictures yielding the dataset of 6,427 subtitle files.

---

[3] See Figure A in the Appendix for a screenshot of https://www.opensubtitles.org website.
[4] See Figure B in the Appendix for a screenshot of https://www.the-numbers.com website.

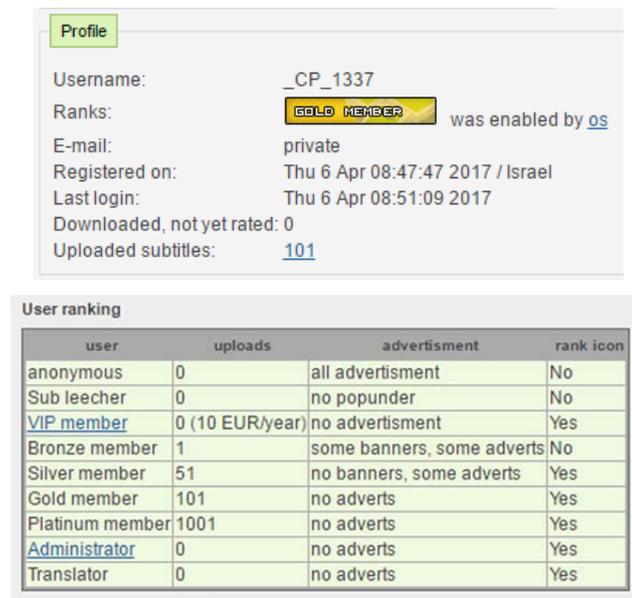

**Figure 2 Screenshot Depicting an Example of Open Subtitles User Ranking and Membership Record**

Finally, the dataset was matched with additional information about motion pictures from IMDb (https://www.imdb.com).[5] This information included: IMDb motion picture ID number; date of release; average IMDb user satisfaction rating from 1 (very bad) to 10 (excellent); critics satisfaction meta score from 0 (very bad) to 100 (excellent); all IMDb genres of the movie (multiple genres were usually listed for each movie on the IMDb website); rating count (number of individual assessments contributing to IMDb rating); number of user reviews; number of critics reviews; number of awards (Oscars and other awards); name of the motion picture director; runtime in minutes; and age appropriateness rating. Matching and further cleansing of the data (removal of duplicates with the same IMDb ID numbers) produced a total final dataset of 6,174 subtitle files.[6]

---
[5] See Figure C in the Appendix for a screenshot of a movie record from the IMDb website.

**3 Methodology and Hypotheses**

We used the resulting filtered dataset of 6,174 movie subtitles to conduct the sentiment analysis of motion pictures. To that end, syuzhet R package was used. Our analysis included the following steps (see Figure 3). First, the emotional arc of each motion picture was calculated by applying the default labelled lexicon developed at the Nebraska Literary Laboratory using cleaned script of each motion picture.[7] To that end, each script was partitioned into sentences and for each sentence the valence was calculated by assigning every word its sentimental value $\sigma \in \{-1, 0, 1\}$, where $\sigma = -1$ referred to emotionally negative terms; $\sigma = 0$ referred to emotionally neutral terms; and $\sigma = 1$ referred to emotionally positive terms according to the lexicon. The resulting sentiment was scaled to fall within the interval $[-1, 1]$. Then the sentiment trajectory was transformed using the Discrete Cosine Transform (DCT). After that, the resulting trajectory was uniformly sub-sampled to have 100 elements so that each motion picture sentiment arc could be represented using the motion picture timing from 0% (beginning of the movie) to 100% (end of the movie).

---

[6] To prepare the subtitles for analysis, we removed time stamp information as well as any special characters not contained in
"abcdefghijklmnopqrstuvwxyzABCDEFGHIJKLMNOPQRSTUVWXYZ'.?!".
[7] See https://github.com/cran/syuzhet/blob/master/README.md for more detail.

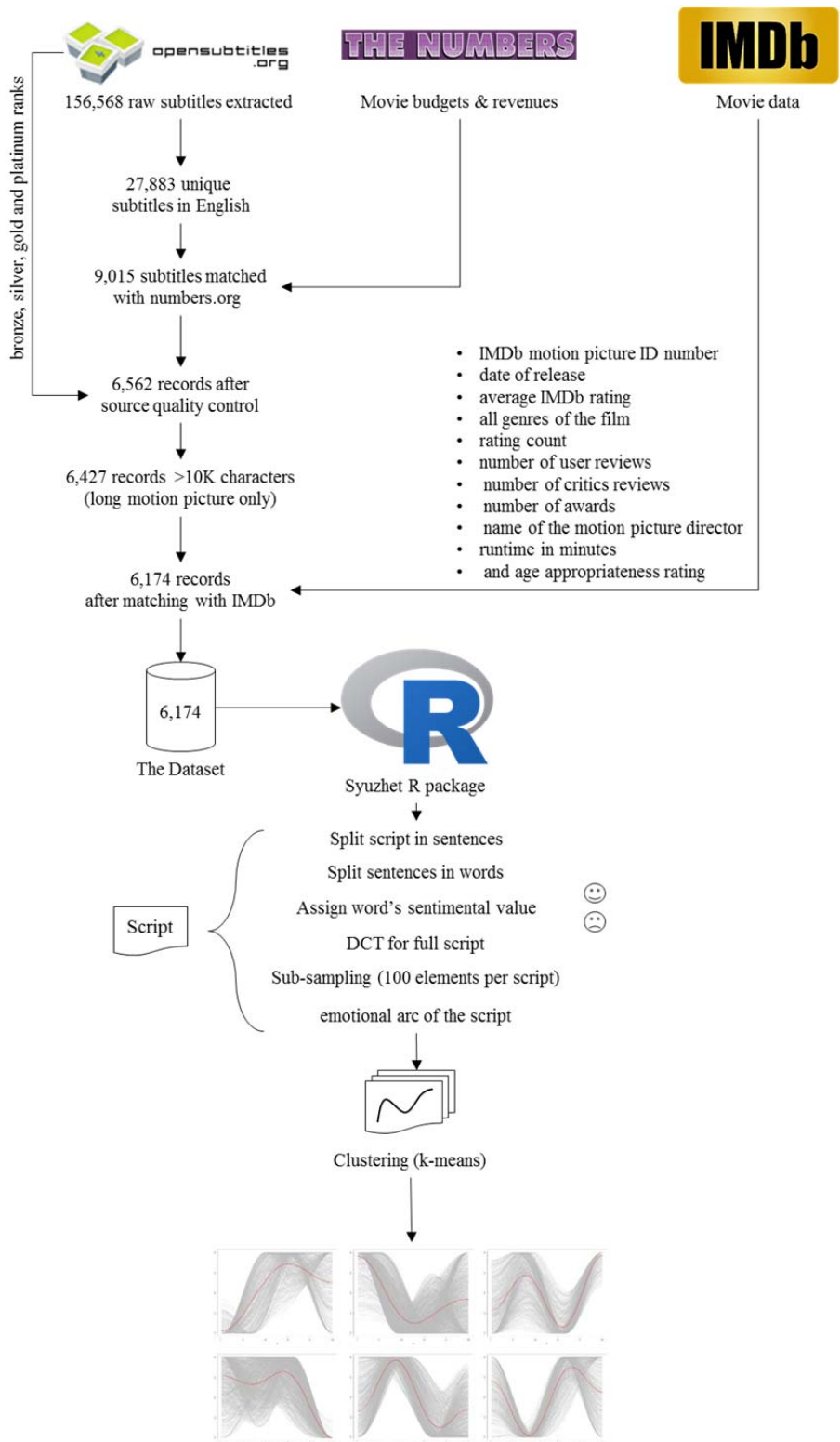

**Figure 3 Steps of the Analysis**

We then accumulated all emotional arcs from motion pictures and applied the following procedure to clustered trajectories. Let functional variable $\chi$ be a random variable taking values in a functional space $\mathcal{E}$. Thus, a functional data set is a sample $\{X_1(t), \ldots, X_N(t)\}_{t=1}^{T}$ drawn from a functional variable $\chi_n$. Here, we represented the sentiment arc associated with a movie $n$ as a realization of $\chi_n$, $\{X_n(1), \ldots, X_n(T)\}$, where $T$ is fixed and $T = 100$.

Clustering on this functional data was carried out using the k-means algorithm in which distances were calculated by approximating the $L_2$ metric:

$$\|X_i(t) - X_j(t)\|^2 = \sqrt{\frac{1}{\omega(t)dt} \int |X_i(t) - X_j(t)|^2 \omega(t)dt}$$

by Simpson's rule, where $\omega(t) \equiv 1$. We used the fda.usc package in R to do the clustering.

Many motion pictures are based on best-selling novels[8]. In part, this may be the case due to risk management: indeed, if a motion picture is based on a popular written content it is more likely to succeed in movie theaters. If this is the case then it is quite likely that movies should generally evoke the same or similar emotions as novels. Hence, we expected to see that, much like novels, motion pictures could be partitioned into the same 6 clusters: *Rags to Riches*, *Riches to Rags*, *Man in a Hole*, *Icarus*, *Cinderella*, and *Oedipus*. Hence, we formulated our first hypothesis as:

**Hypothesis 1: Emotional arcs generated by movies fit the same 6 clusters as novels.**

Reagan et al. (2016) found that *Icarus*, *Oedipus*, and *Man in a Hole* produce more successful novels when success is measured by the number of downloads. We expect that the

---

[8] For a recent account of how books translate into movies see
https://www.theverge.com/2017/1/26/14326356/hollywood-movie-book-adaptations-2017-expanse-game-of-thrones

same three emotional arcs' clusters will perform well in the movie theaters. Specifically, our second hypothesis is:

**Hypothesis 2: Similarly to novels, motion picture emotional arcs resembling *Icarus*, *Oedipus*, and *Man in a Hole* shapes are associated with more successful movies.**

Our dataset allowed us to use several measures of movie success. Specifically, we considered revenue figures, movie awards, as well as satisfaction indicators to assess the success of the motion picture. Additionally, we were also able to explore how emotional arcs in conjunction with other indicators affect movie success. Specifically, we considered how genres combined with emotional arc clusters affected success variables. Budget estimates gave us an opportunity to conduct a robustness check of our results.

## 4 Results

In this section we test our hypotheses and explore how robust our results are. We find that similarly to novels (see Reagan, 2016), all analyzed movie scripts can be partitioned to fit 6 major emotional trajectories (clusters) where each trajectory is obtained using the clustering procedure described in Section 3[9]. This confirms our Hypothesis 1. Figure 4 shows all 6 clusters of emotional trajectories and provides examples of films which fall within each cluster.

---

[9] To check for robustness of our results, we have conducted clustering procedure using 4, 6, 8, 10, and 12 clusters. Our analysis shows that 6 is the optimal number of clusters as <6 clusters result in imprecise fitting of the general pattern functions and >6 clusters produce similar clusters which are hard to distinguish looking at the resulting functional forms. Results of the robustness check clustering are available from the corresponding author upon request.

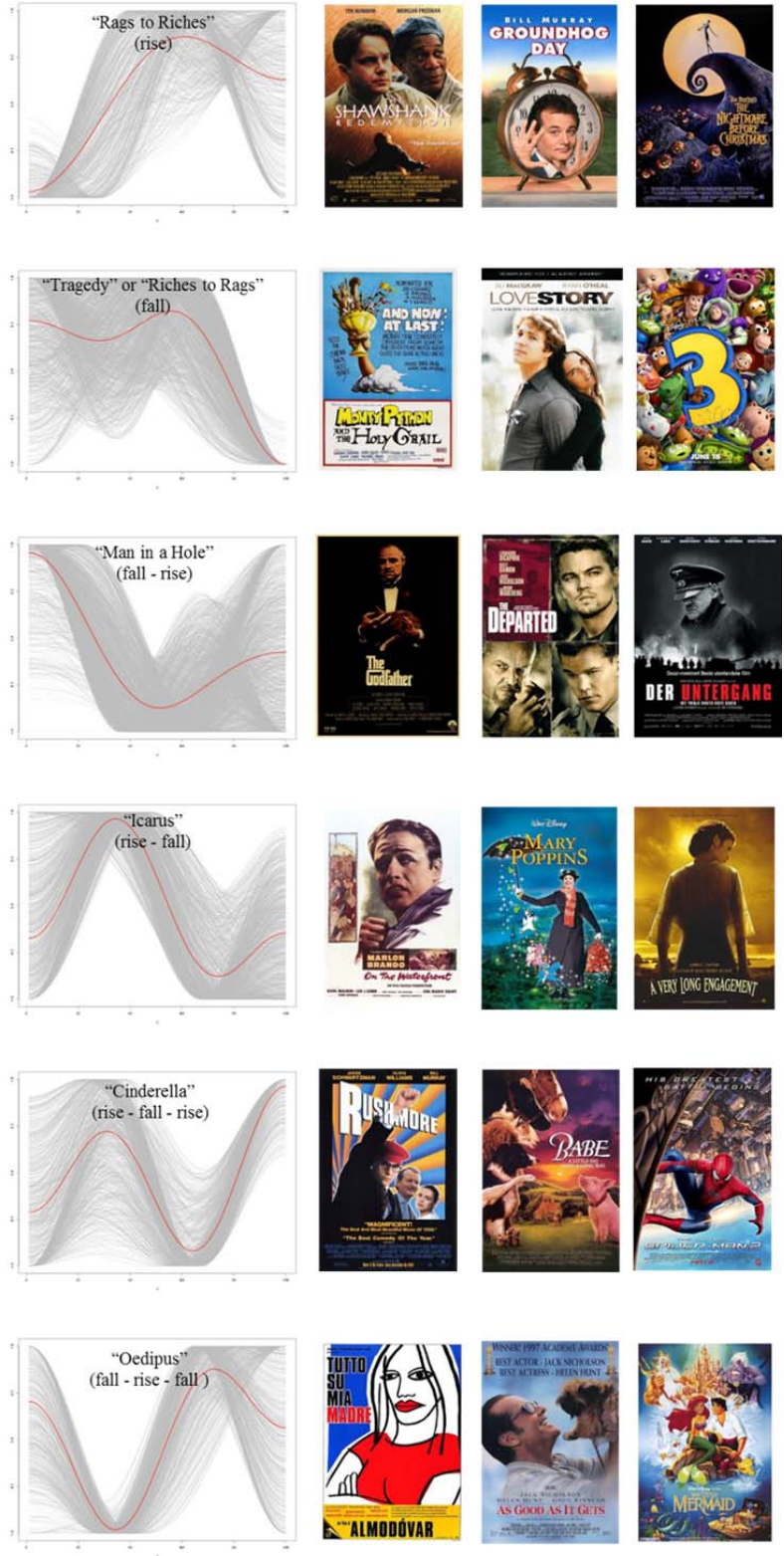

**Figure 4 Six Emotional Trajectories of Movies**

**4.1 Emotional Arcs and Success of Motion Pictures**

Our resulting filtered dataset of 6,174 movies consists of 632 movies in the *Rags to Riches* cluster; 1402 movies in the *Riches to Rags* cluster; 1598 movies in the *Man in a Hole* cluster; 1113 in the *Icarus* cluster; 804 movies in the *Cinderella* cluster; and 625 movies in the *Oedipus* cluster (see Table 1 for summary statistics). Therefore, each cluster contains at least 625 movies. According to Table 1, movies are relatively balanced in terms of length with average run times between 108 and 110 minutes.

In order to compare the success of movies in each emotional trajectory cluster we first consider domestic gross revenue as a success indicator. We initially use this variable because we could not find worldwide gross revenue for all movies in our dataset, yet domestic gross revenue was available for all 6,174 movies.[10] Table 1 shows that top three clusters in terms of mean gross domestic revenue are *Man in a Hole* (earning $37.48 million on average); *Cinderella* (with $33.63 million mean revenue); and *Oedipus* (yielding $31.44 million on average). Notably, two of the three top earning emotional trajectories in our analysis coincide with those found by Reagan et al (2016). Specifically, while *Man in a Hole* and *Oedipus* emotional trajectories are associated with the most downloaded e-books as well as with the highest revenue-generating movies, *Cinderella* trajectory outperforms *Icarus* in movie theaters. This may indicate that people's desired emotions depend on the time length of their experience. Specifically, it is safe to assume that the same story is experienced in more condensed time when one watches a movie compared to when one reads a book.

---

[10] We use worldwide gross revenue variable in later subsections and show that our results are essentially the same when we consider domestic gross revenue and worldwide gross revenue.

Table 1 Summary Statistics

|  | Rags to Riches N=632 | Riches to Rags N=1,402 | Man in a Hole N=1,598 | Icarus N=1,113 | Cinderella N=804 | Oedipus N=625 | Total N=6,174 |
|---|---|---|---|---|---|---|---|
| Domestic gross revenue | 29.71<br>8.56<br>49.89 | 29.94<br>7.84<br>60.80 | 37.48<br>13.45<br>63.70 | 30.57<br>7.92<br>53.93 | 33.63<br>10.84<br>56.78 | 31.44<br>9.07<br>58.64 | 32.61<br>9.67<br>58.68 |
| IMDb user rating | 6.64<br>6.70<br>0.97 | 6.52<br>6.60<br>0.98 | 6.45<br>6.50<br>0.99 | 6.54<br>6.70<br>0.98 | 6.51<br>6.70<br>0.99 | 6.52<br>6.60<br>0.94 | 6.52<br>6.60<br>0.98 |
| IMDb meta score | 57.12<br>57.00<br>18.26 | 57.62<br>59.00<br>17.73 | 55.44<br>55.50<br>17.80 | 56.27<br>57.00<br>17.98 | 56.42<br>56.00<br>17.83 | 56.24<br>57.00<br>17.41 | 56.45<br>57.00<br>17.84 |
| Rating count | 67,766<br>20,388<br>135,458 | 67,118<br>22,490<br>140,525 | 78,166<br>27,490<br>139,245 | 67,385<br>21,152<br>124,867 | 70,177<br>22,495<br>123,657 | 58,056<br>20,299<br>103,016 | 69,573<br>22,908<br>131,450 |
| User reviews | 208<br>98<br>363 | 226<br>111<br>359 | 239<br>129<br>351 | 208<br>111<br>306 | 225<br>108<br>343 | 186<br>102<br>279 | 220<br>112<br>339 |
| Critics' reviews | 115<br>81<br>110 | 121<br>90<br>112 | 133<br>97<br>120 | 117<br>83<br>112 | 122<br>87<br>113 | 106<br>71<br>103 | 121<br>86<br>113 |
| Oscars Won | 0.35<br>0.00<br>0.99 | 0.33<br>0.00<br>1.05 | 0.34<br>0.00<br>0.97 | 0.33<br>0.00<br>0.96 | 0.28<br>0.00<br>0.91 | 0.34<br>0.00<br>1.02 | 0.33<br>0.00<br>0.99 |
| Other awards | 6.72<br>2.00<br>14.90 | 5.94<br>2.00<br>13.53 | 5.92<br>2.00<br>15.70 | 5.92<br>2.00<br>11.88 | 6.04<br>2.00<br>15.57 | 4.88<br>1.00<br>9.17 | 5.92<br>2.00<br>13.92 |
| Other awards nominations | 11.22<br>4.00<br>23.17 | 10.39<br>3.50<br>21.04 | 11.37<br>4.00<br>25.53 | 10.06<br>4.00<br>18.98 | 10.77<br>4.00<br>23.75 | 9.20<br>3.00<br>17.35 | 10.60<br>4.00<br>22.21 |
| Movie length | 110<br>106<br>22 | 108<br>104<br>21 | 108<br>104<br>22 | 110<br>106<br>21 | 108<br>104<br>21 | 108<br>103<br>20 | 108<br>104<br>21 |

Note: Each cell shows mean (top row), median (middle row) and standard deviation (bottom row). Gross revenue is measured in million US dollars.

Specifically, movies in our dataset last on average 108 minutes while reading a book with a similar story would take an average reader many hours if not days. In other words, consumption time for a book is greater than that for a movie. Consequently, one reason why *Icarus* movies do not do as well as *Icarus* books could be that in a time-limited environment people do not want to experience emotional fall which is not followed by an equivalent or nearly equivalent emotional rise.

However, people are quite happy to experience such a dramatic fall during a larger period of time when the intensity of emotional fall is diffused (i.e., when reading a book). In contrast, *Cinderella* emotional trajectory provides a noticeable emotional rise towards the end of the story despite the emotional fall in the middle of the movie. This emotional rise may be more desirable for the viewers of the movies compared to the readers of the books.

At the first glance, if we consider mean values of the gross domestic revenue as a proxy of success, two of three clusters of emotional trajectories are the same for movies and books. Yet, are movies in these three clusters earning <u>statistically significantly more</u> than movies in other clusters? We conducted a series of OLS regressions with domestic gross revenue as a dependent variable and dummies for each of the emotional trajectories to understand whether obtained differences in revenues are statistically significant.[11] Our results show that only one cluster – *Man in a Hole* – produces statistically significantly higher gross domestic revenue compared with other clusters. Moreover, in a regression analysis *Oedipus* cluster reveals negative (though not statistically significant) correlation with gross domestic revenue. As shown in Table 2, the effect of *Man in a Hole* cluster is high (the coefficient is equal to 6.5613 suggesting that producing a movie with *Man in a Hole* emotional arc is equivalent to the mean increase in gross domestic revenue of over $6 million), positive, and significant at 0.1% level.

---

[11] Unfortunately, Reagan et al. (2016) do not provide statistical significance levels for their results which makes it difficult for us to compare our findings to those reported in their paper.

Four emotional trajectory clusters: *Cinderella*, *Oedipus*, *Icarus*, and *Rags to Riches* do not reveal statistically significant results.

Interestingly, the *Riches to Rags* cluster shows a negative and statistically significant correlation with domestic gross revenue. The effect is quite large (the coefficient of -3.4599 indicates that producing a film with *Riches to Rags* emotional arc is equivalent to the mean decline in domestic revenue of more than $3 million).

Table 2 reports several interesting results regarding other success indicators. Specifically, even though *Man in a Hole* produces statistically significantly higher gross domestic revenue than any other emotional arc, the IMDb ratings associated with this emotional arc are negative and significant. The effect of the arc on IMDb user rating is rather small yet significant. According to Table 1 IMDb user ratings for all emotional trajectory clusters are very similar: 4 of 6 clusters have an average rating close to 6.5; *Man in a Hole* has a mean rating of 6.45 and *Rags to Riches* has an average rating of 6.64. Regression results reported in Table 2 show that there is a positive and statistically significant correlation between movies in the *Rags to Riches* cluster and IMDb rating although the effect is small.

The *Man in a Hole* cluster is also not associated with high critics scores on IMDb. Specifically, there is a negative and statistically significant correlation between IMDb critics meta score and *Man in a Hole* cluster. At the same time, critics meta score is positively correlated with *Riches to Rags* cluster which tend to be associated with low revenues. These results suggest that critics tend to prefer stern movies (possibly with an unhappy ending) and these movies tend to be less successful in generating revenue.

**Table 2 Results of the Series of OLS Regressions with Emotional Arcs as Independent Variables**

| Dependent variable | Rags to Riches=1, 0 otherwise | Riches to Rags=1, 0 otherwise | Man in a Hole=1, 0 otherwise | Icarus=1, 0 otherwise | Cinderella=1, 0 otherwise | Oedipus=1, 0 otherwise |
|---|---|---|---|---|---|---|
| Domestic gross revenue | -3.2333 (2.4636) | -3.4599* (1.7823) | 6.5613*** (1.7032) | -2.4914 (1.9427) | 1.1690 (2.2192) | -1.3031 (2.4761) |
| IMDb user rating | 0.1406*** (0.0412) | 0.0020 (0.0298) | -0.0910*** (0.0285) | 0.0307 (0.0325) | -0.0063 (0.0371) | 0.0040 (0.0414) |
| IMDb meta score | 0.7475 (0.8606) | 1.5060* (0.6297) | -1.3815* (0.5893) | -0.2189 (0.6847) | -0.0319 (0.7722) | -0.2369 (0.8763) |
| Rating count | -2013.03 (5519.30) | -3175.87 (3993.32) | 11593.47** (3817.02) | -2669.58 (4352.13) | 694.92 (4971.24) | -12814.16* (5544.28) |
| User reviews | -13.2979 (14.2238) | 7.8058 (10.2919) | 25.5148** (9.8395) | -14.6074 (11.2154) | 5.3835 (12.8120) | -38.4024** (14.2869) |
| Critics' reviews | -6.9008 (4.7592) | -0.0751 (3.4441) | 15.0109*** (3.2888) | -5.0277 (3.7529) | 0.9112 (4.2873) | -17.5043*** (4.7783) |
| Oscars won | 0.0221 (0.0414) | -0.0009 (0.0299) | 0.0135 (0.0286) | 0.0004 (0.0326) | -0.0496 (0.0373) | 0.0121 (0.0416) |
| Other awards | 0.8904 (0.5842) | 0.0340 (0.4227) | 0.00619 (0.4044) | -0.0005 (0.4607) | 0.1381 (0.5262) | -1.1491* (0.5870) |
| Other awards nominations | 0.6926 (0.9324) | -0.2737 (0.6746) | 1.0482 (0.6452) | -0.6600 (0.7352) | 0.1991 (0.8398) | -1.5576 † (0.9368) |

Note: Each cell reports the OLS regression coefficient followed by a standard error in square brackets. † - significant at 10% level – $p<0.1$; * - significant at 5% level – $p<0.05$; ** - significant at 1% level – $p<0.01$; *** - significant at 0.1% level – $p<0.001$.

Why does the *Man in a Hole* emotional arc produce high revenue but does not generate high user and critics' ratings on IMDb? There could be several reasons for this: (1) people are more likely to leave feedback (rating or review) if they did not have a good experience so it could be that there is some bias in the IMDb satisfaction scores which, generally, are lower than the average viewers' attitude or (2) IMDb scores are provided by a different audience than that which primarily contributes to the movie revenue, etc. More insight into the difference between IMDb ratings and gross domestic revenue is provided by further variables capturing the number of people leaving ratings and reviews. All three variables that capture the level of activity on IMDb – rating count, the number of user reviews, and the number of critics' reviews – are positively and significantly correlated with the *Man in a Hole* emotional trajectory. If we assume that the mean IMDb user rating and the IMDb meta score could be taken as a proxy of viewers' and critics satisfaction respectively, our results may suggest that highest earning movies are not necessarily the ones that are liked by the audience, but rather are those that attract the most attention. In other words, the *Man in a Hole* emotional trajectory does not produce the "most liked" movies, but generates the most "talked about" movies.

To verify the relations between different proxies of success used in our analysis, we conduct a clustered OLS regression analysis (where standard errors are clustered at the level of each emotional arc) with gross domestic income as a dependent variable and IMDb success indicators as independent variables. Results of this analysis are reported in Table 3. Our findings summarized in Table 3 confirm our conjecture that high IMDb ratings are not associated with the highest revenue. Specifically, while user ratings (satisfaction indicators) are generally negatively correlated with the gross domestic revenue, popularity indicators (number of ratings, number of user and critics' reviews) are positively correlated with the gross domestic revenue. For robustness, we have also conducted the same analysis using worldwide revenue for a reduced sample of movies (3,051 observations in our dataset contained information on worldwide revenues). Table 3 shows that results of the OLS clustered regression with worldwide revenue as a dependent variable essentially repeat those with gross domestic revenue as a dependent variable.

Table 3 Correlations between Success Variables:

Clustered OLS Regression Results

|  | Dependent variable | |
|---|---|---|
| Independent variables | Domestic gross revenue | Gross revenue worldwide |
| IMDb user rating | -7.8031** (1.3479) | -23.5586*** (3.7306) |
| IMDb meta score | 0.0124 (0.0557) | -0.0152 (0.1288) |
| Rating count | 0.0002*** (0.0000) | 0.0005*** (0.0001) |
| User reviews | 0.0301*** (0.0046) | 0.0373† (0.0162) |
| Critics' reviews | 0.0988*** (0.0135) | 0.3653 *** (0.0230) |
| Oscars won | 7.6432** (1.8880) | 15.2713† (6.3472) |
| Other awards | -0.3665** (0.0926) | -0.4353 (0.4167) |
| Other awards nominations | -0.0993 (0.0522) | -0.4436 (0.2389) |
| Constant | 50.1448*** (6.3387) | 134.9916** (23.0305) |
| $R^2$ | 0.4738 | 0.4076 |
| N | 6,147 | 3,051 |

Notes: † - significant at 10% level - $p<0.1$;

* - significant at 5% level – $p<0.05$;

** - significant at 1% level – $p<0.01$;

*** - significant at 0.1% level – $p<0.001$.

Table 2 also shows that the *Oedipus* cluster does not generate many ratings and reviews compared to other clusters. Despite being one of the top 3 earning arcs according to the average indicators reported in Table 1, the *Oedipus* cluster is negatively correlated with gross domestic revenue, though this correlation is not statistically significant according to Table 2. Nevertheless, this cluster also produces a negative correlation with non-Oscar awards and non-Oscar award nominations. Specifically, *Oedipus* movies are less likely to be nominated for non-Oscar awards, and less likely to receive them than any other cluster (see Table 2). Interestingly, according to Table 3, Oscars are generally associated with higher domestic and worldwide revenue. However, this could be due to increased popularity following an Oscar award as well as the fact that production companies often carefully select release dates for Oscar-nominated movies.[12]

**4.2 Emotional Arcs and Movie Budgets**

So far, we have established that the *Man in a Hole* emotional trajectory generates the highest gross domestic revenue which partially confirms our Hypothesis 2. We also found that (based on assumption that IMDb rating indeed capture viewer satisfaction rates) this emotional trajectory is top earning not because it produces the most "liked" content but because movies in this cluster attract most viewers attention. We now turn to the robustness check of our results and explore whether and how production budgets affect revenues.

Motion pictures are expensive to produce and it is important to understand whether and to what extent high revenue is associated with the level of initial investment in movie production. To explore this issue, we look at the estimated production budgets obtained from

---

[12] See, e.g., https://www.theatlantic.com/entertainment/archive/2013/01/release-dates-oscars/319514/ for more detail.

https://www.the-numbers.com repository for a subsample of our dataset. Specifically, for 3,051 movies we have budget information. It is important to note that the repository only provides budget estimates. This is due to the fact that budget figures are usually a part of the production commercial secret. Specifically, https://www.the-numbers.com provides the following statement about movie production budget figures: "*Budget numbers for movies can be both difficult to find and unreliable. Studios and film-makers often try to keep the information secret and will use accounting tricks to inflate or reduce announced budgets. This chart shows the budget of every film in our database, where we have it. The data we have is, to the best of our knowledge, accurate but there are gaps and disputed figures.*" With this limitation in mind we first summarize statistics for a subsample of movies in our dataset for which we have gross domestic revenue, worldwide revenue, as well as estimated budgets (see Table 4).

As we can see from Table 4, the *Man in a Hole* emotional trajectory cluster generates the highest revenue not only according to the values obtained from our total sample of 6,174 movies, but also according to the numbers obtained using a subsample of movies with budget estimates (3,051 movies). This is true for both the gross domestic revenue as well as for the worldwide revenue. A series of OLS regressions reveal that *Man in a Hole* is the only emotional trajectory which produces statistically significant results showing that it is more financially successful than any other emotional arc using a subsample of data with budgets. This is the case for gross domestic revenue (the coefficient is equal to 5.217438 with standard error of 2.713389 and a significance level of $p=0.055$); as well as for the worldwide revenue (the coefficient is equal to 12.02102 with standard error of 7.043771 and a significance level of $p=0.088$). In other words, our result that *Man in a Hole* is generating the highest revenue is confirmed for both gross domestic revenue and worldwide revenue using a smaller sample of data though (unsurprisingly) the statistical significance level decreases for a smaller sample (both regression coefficients are significant at 10% level).

**Table 4 Estimated Budgets, Gross Domestic Revenue, and Worldwide Revenue for a Subsample of 3,051 Motion Pictures**

| Clustered emotional trajectory | Subsample (# of motion pictures) | Gross domestic revenue N=6,174 | Subsample with budget estimates | Budget estimate N=3,051 | Gross domestic revenue N=3,051 | Worldwide revenue N=3,051 |
|---|---|---|---|---|---|---|
| Rags to Riches | 632 | 29.71 8.56 49.89 | 306 | 36.27 25.00 37.90 | 48.64 31.22 59.75 | 101.39 48.60 140.29 |
| Riches to Rags | 1,402 | 29.94 7.84 60.80 | 638 | 35.94 20.00 43.38 | 49.92 25.70 76.73 | 107.28 42.10 206.28 |
| Man in a Hole | 1,598 | 37.48 13.45 63.70 | 874 | 40.50 28,00 42.15 | 54.89 33.12 67.93 | 118.76 58.90 172.49 |
| Icarus | 1,113 | 30.57 7.92 53.93 | 534 | 35.74 22,00 39.26 | 48.99 27.88 62.95 | 103.28 48.86 168.51 |
| Cinderella | 804 | 33.63 10.84 56.78 | 407 | 39.04 24.00 42.90 | 51.70 26.54 67.89 | 116.83 43.60 178.81 |
| Oedipus | 625 | 31.44 9.07 58.64 | 292 | 38.19 24.50 44.82 | 48.65 27.95 62.74 | 103.40 43.49 155.98 |
| Total | 6,174 | 32.61 9.67 58.68 | 3,051 | 37.87 25.00 41.89 | 51.17 29.08 67.79 | 110.18 48.06 175.96 |

Notes: Each cell reporting revenue and budget numbers shows mean value in the top row, median value in the middle row, and standard deviation in the bottom row.

Table 4 also reveals that the *Man in a Hole* movies are associated with the highest average estimated budget. Specifically, for our subsample of 3,051 movies with budget information, *Man in a Hole* movies on average cost $40.5 million to produce (and earn on

average $54.9 million), while *Cinderella* movies have a mean estimated production budget of $39 million (and earn on average $51.7 million), *Oedipus* movies cost $38.2 million (and earn $48.7 million); *Rags to Riches* - $36.3 million (and earn $48.6 million); and *Icarus* – $35.7 million (earning almost $49 million). Does it mean that the *Man in a Hole* emotional trajectory simply requires more investment and this drives higher revenue? If this is the case, then we should observe (i) that budgets for *Man in a Hole* movies are significantly higher than those for movies within other emotional arcs; and (ii) that there is a higher dependency between budget numbers and the *Man in a Hole* cluster compared to all other clusters. To test our conjecture (i), we first conduct a series of pairwise non-parametric comparisons between *Man in a Hole* cluster budgets and budgets of all other clusters. A series of Mann-Whitney Wilcoxon test (comparing budget means) show that *Man in a Hole* movie budgets are not statistically significantly different from budgets of the *Rags to Riches* cluster ($p>0.10$), *Cinderella* cluster ($p>0.18$), and *Oedipus* cluster ($p>0.16$) but higher than average budgets of *Riches to Rags* ($p<0.001$) and *Icarus* movies ($p<0.005$). Furthermore, the Kolmogorov-Smirnov test (comparing distributions of budgets) also shows no difference between *Man in a Hole* and *Rags to Riches* ($p>0.27$), *Cinderella* ($p>0.12$), and *Oedipus* ($p>0.31$) budgets and significant difference between *Man in a Hole* and *Riches to Rags* ($p<0.01$) and *Icarus* ($p<0.05$). If budget indeed was the main determinant of the revenue, we should have seen *Rags to Riches*, *Cinderella*, and *Oedipus* (as *Man in a Hole*) generate statistically significantly greater revenues compared to *Riches to Rags* and *Icarus*. Yet, this is not the case.

To test our conjecture (ii), we look at the relation between budgets and revenues for each emotional arc (see Figure 5). A series of OLS regressions with gross domestic revenue (Figure 5 (a)) and worldwide revenue (Figure 5 (b)) as dependent variables and estimated budget as an explanatory variable shows that movie budgets are positively correlated with revenues for all emotional arcs.

(a) Gross Domestic Revenue and Budgets

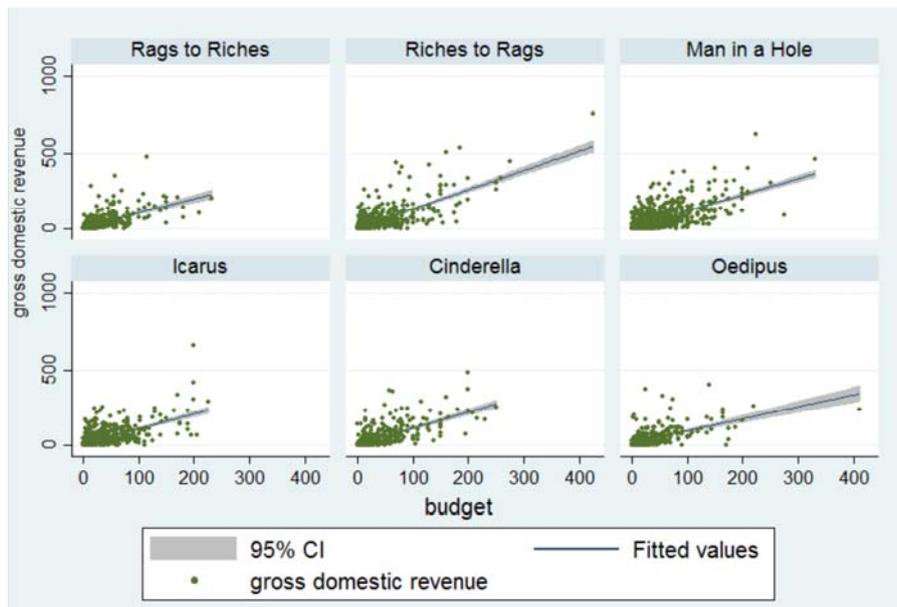

(b) Worldwide Revenue and Budgets

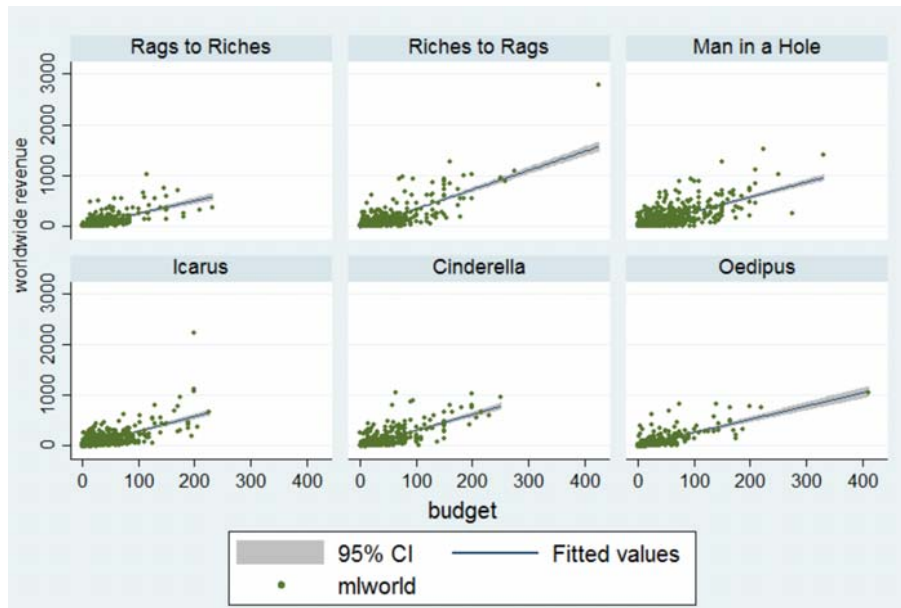

**Figure 5 Correlations between Movie Production Budgets and Revenues by Emotional Arc**

When we use gross domestic revenue as a dependent variable, this relationship is highly statistically significant (at 0.1% level) for all clusters. Furthermore for 5 clusters: *Rags to Riches* (regression coefficient 0.91); *Riches to Rags* (regression coefficient 1.27); *Man in a Hole* (regression coefficient 1.07); *Icarus* (regression coefficient 1.06) regression coefficients are similar and close to 1 (meaning that a $1 million increase in budget usually leads to approximately $1 million increase in revenue). Only for the *Oedipus* emotional arc do we observe a slightly lower regression coefficient of 0.80. Furthermore, one of the least financially successful arcs – Riches to Rags – has the highest regression coefficient. Results obtained for the gross domestic revenue are confirmed for the worldwide revenue (see Figure 5 (b)). For worldwide revenue, the relationship between budgets and revenues for all emotional arcs are positive and highly significant at 0.1% level; and coefficients range between 2.59 (lowest coefficient) for the *Oedipus* cluster and 3.76 for the *Riches to Rags* cluster (highest coefficient). This means that even though budget plays an important role in movie production and contributes to the motion picture's subsequent financial success, the *Man in a Hole* emotional arc does not have a higher dependency on budget than other emotional arcs. Therefore, heterogeneity in production budgets cannot explain the *Man in a Hole* relative financial success compared to other arcs.

So far we have established that the financial success of the *Man in a Hole* emotional arc cannot be explained by higher financial investment. We now explore whether and to what extent the compound effect of budget and emotional arc contributes to motion picture revenue. This allows us to understand whether Man in a Hole financial success is driven by movies falling within a particular budget category.

We partition movies into 8 categories according to the production budget variable: (1) movies with budgets of up to $1 million (N=107); (2) movies with budgets between over $1 million and $5 million (N=346); (3) movies with budgets between over $5 million and $10 million (N=339); (4) movies with budgets between over $10 million and $20 million (N=615); (5) movies with budgets of over $20 million and $30 million (N=399); (6) movies with budgets between over $30 million and $50 million (N= 512); (7) movies with budgets between over $50 million and $100 million (N=518); and (8) movies with budgets over $100 million (N=215). We then conduct a series of OLS regressions for movies in each emotional arc falling within each of the budget categories. Gross domestic revenue was used as a dependent variable[13] and emotional clusters – as explanatory variables. Table 5 summarizes our results in a simple heat map.[14]

Table 5 shows that the *Man in a Hole* emotional arc produces higher revenue than any other arc; but this financial success is not due to movies falling within any particular budget category. Even though overall *Riches to Rags* is the least financially successful arc, movies in this cluster seem to generate statistically significantly high revenue when they are in a high budget category (over $100 million). This may explain the financial success of large historical drama productions such as *The Last Samurai* or survival epics like *Life of Pi*.

Table 5 also shows that the *Icarus* type of movies tend to succeed when they are low to medium budget productions (i.e., productions of under $1 million and productions between over $5 and $10 million) and fail when they require large financial investment (movies with budgets between over $50 million and $100 million). The Table also reveals that *Cinderella*

---

[13] In one of the OLS regressions (captured in a penultimate row of Table 5) we have checked the robustness of our results using worldwide revenue as a dependent variable.
[14] Detailed results are reported in Table A in the Appendix.

movies with budgets between over $1 million and $5 million as well as *Oedipus* motion pictures with budgets between over $30 million and $50 million tend to be less financial successful than movies in other categories.

**Table 5 Compound Effect of Emotional Arcs and Budgets**

| | Arc type | | | | | |
|---|---|---|---|---|---|---|
| Budget (millions) | Rags to Riches | Riches to Rags | Man in a Hole | Icarus | Cinderella | Oedipus |
| [0,1] | -1 | -1 | 1 | 2 | 1 | -1 |
| ]1; 5] | -1 | 1 | 1 | 1 | -2 | 1 |
| ]5; 10] | -1 | 1 | -1 | 3 | -1 | -1 |
| ]10; 20] | 1 | -1 | -1 | 1 | 1 | -1 |
| ]20; 30] | 1 | -1 | -1 | 1 | -1 | 2 |
| ]30; 50] | 1 | -1 | 1 | 1 | -1 | -2 |
| ]50;100] | -1 | -1 | 1 | -3 | 1 | 1 |
| ]100;∞[ | -1 | 3 | -1 | -1 | -1 | -1 |
| All budget (gross domestic revenue) | -1 | -1 | 3 | -1 | 1 | -1 |
| All budget (worldwide revenue) | -1 | -1 | 2 | -1 | 1 | -1 |
| All data (gross domestic revenue) | -1 | -2 | 5 | -1 | 1 | -1 |

Note: 1 and -1 - not significant with positive and negative effect respectively
2 and -2 - significant at 10% with positive and negative effect respectively
3 and -3 - significant at 5% with positive and negative effect respectively
4 and -4 - significant at 1% with positive and negative effect respectively
5 and -5 - significant at 0.1% with positive and negative effect respectively
n.a. – no observations

### 4.3 Emotional Arcs and Genres

In the previous subsection we explored the impact of production budgets on movies' financial revenues. Yet, other factors may influence movie success. One such factor is movie genre. In this subsection we investigate whether and how movie genres influence revenue. To that end, we look at the compound effects of movie genres and emotional arcs by conducting a series of OLS regressions with gross domestic revenue as a dependent variable[15] and emotional arc clusters as explanatory variables for all combinations of genre and emotional arc in our sample. Genre information is obtained from the movie description on the IMDb website which lists 22 possible genres: Action, Horror, SciFi, Mystery, Thriller, Animation, Drama, Adventure, Fantasy, Crime, Comedy, Romance, Family, Biography, Sport, Music, War, Western, History, Musical, Film Noir, and News. It is important to note that most motion pictures which appear on the IMDb website are characterized by more than one genre. Our resulting dataset consisted of 1,201 Action movies; 564 Horror movies; 597 SciFi movies; 594 Mystery movies; 1,726 Thrillers; 268 Animations; 3,757 Dramas; 961 Adventure movies; 644 Fantasy movies; 1,219 Crime movies; 2,403 Comedies; 1,710 Romance movies; 584 Family movies; 386 Biographies; 187 Sport-themed movies; 262 Music-related movies; 281 War-themed; 88 Westerns; 234 History movies; 170 Musicals; and 8 Film Noir movies. Even though News was listed on IMDb as a genre, there were no movies in that category. To make use of all the available information, we constructed dummies for all genres and then looked at the revenues of movies falling within each genre category separately. Table 6 summarizes our results using a simple heat map.[16]

Table 6 shows that for most genres, the *Man in a Hole* emotional arc produces high revenue and for SciFi, Mystery, Thriller, Animation, Adventure, Fantasy, Comedy, and Family movies this result is statistically significant. *Riches to Rags* motion pictures succeed financially if they fall within Biography, Music, War, or History genres; yet fail if they are SciFi motion pictures, Mysteries, or Thrillers.

---

[15] Results for the worldwide revenue are essentially the same. We report gross domestic revenue results to make use of our entire sample of 6,147 movies as worldwide revenue is only available for 3,051 movies in our dataset. Worldwide revenue results are available from the corresponding author upon request.
[16] Detailed results are reported in Table B in the Appendix.

**Table 6 Compound Effect of Movie Genre and Emotional Arc on Gross Domestic Revenue**

Emotional Arc

| IMDb Genre | Rags to Riches | Riches to Rags | Man in a Hole | Icarus | Cinderella | Oedipus |
|---|---|---|---|---|---|---|
| Action | -1 | -1 | 1 | -1 | -1 | -1 |
| Horror | -1 | 3 | 1 | -1 | -1 | -1 |
| SciFi | -2 | 1 | 4 | -1 | -1 | -1 |
| Mystery | -3 | 1 | 3 | 1 | -1 | -1 |
| Thriller | -3 | 3 | 4 | -1 | -1 | -4 |
| Animation | -1 | -2 | 3 | -2 | 1 | 1 |
| Drama | 1 | -1 | 1 | 1 | -1 | 1 |
| Adventure | -1 | -2 | 4 | -1 | 1 | -1 |
| Fantasy | -1 | -1 | 4 | -3 | 1 | -1 |
| Crime | -1 | 1 | 1 | 1 | -3 | -2 |
| Comedy | -1 | -5 | 5 | -2 | 3 | 1 |
| Romance | 1 | -1 | -1 | -1 | 1 | 2 |
| Family | -1 | -1 | 4 | -3 | 2 | -1 |
| Biography | 5 | -1 | -1 | -1 | 1 | -1 |
| Sport | -1 | 1 | -1 | -1 | -1 | 5 |
| Music | 3 | -3 | -1 | -1 | 1 | 1 |
| War | 2 | 1 | -1 | 1 | -2 | -1 |
| Western | -1 | 2 | -1 | -1 | -1 | 1 |
| History | 3 | -2 | 1 | -1 | -1 | 1 |
| Musical | -1 | 1 | 1 | -1 | -1 | -1 |
| Film Noir | -1 | 1 | -1 | 1 | -1 | -1 |
| News | n.a. | n.a. | n.a. | n.a. | n.a. | n.a. |

Note: 1 and -1 - not significant with positive and negative effect respectively
2 and -2 - significant at 10% with positive and negative effect respectively
3 and -3 - significant at 5% with positive and negative effect respectively
4 and -4 - significant at 1% with positive and negative effect respectively
5 and -5 - significant at 0.1% with positive and negative effect respectively
n.a. – no observations

*Riches to Rags* Horrors, Westerns and Thrillers tend to achieve high revenue while *Riches to Rags* Animations, Adventures, Comedies, Music-themed, and Historical movies tend to have low revenues. *Icarus* movies tend to generate low revenues irrespective of the genre with *Icarus* Animations, Fantasies, Comedies, and Family movies being especially low earning. *Cinderella* motion pictures tend to achieve high revenues as Comedies and Family movies but low revenues as Crime and War-themed movies. Finally, *Oedipus* motion pictures do well as Romance and Sport-themed movies but tend to fail as Thrillers and Crime-themed movies.

These results allow us to explore an extra dimension of the motion pictures success. Our findings show that while *Rags to Riches*, *Riches to Rags*, *Cinderella*, and *Oedipus* movies may produce different revenues dependent on the genre; *Icarus* motion pictures tend to be financially unsuccessful irrespective of the genre and *Man in a Hole* movies, on the contrary, tend to generate high revenues across the majority of genres. Clearly, there is some heterogeneity among *emotional arcs - genres* combinations revealing that, in principle, many emotional arcs (with the exception of *Icarus*) may be associated with financially successful movies. Yet, it is also clear that the *Man in a Hole* emotional arc tends to financially outperform other arcs in the majority of genre variations.

## 5 Conclusion

Recent advances in data science allow us to better understand emotions and use this knowledge to predict viewers' preferences more accurately. In our recent study we use data science NLP methodology to explore whether and to what extent emotions shape consumer preferences for media and entertainment content. We find that all analyzed emotional arcs from thousands of motion picture scripts can be partitioned into 6 major emotional trajectories: *Rags to Riches*, *Riches to Rags*, *Man in a Hole*, *Icarus*, *Cinderella*, and *Oedipus*. We also find that one of these trajectories – *Man in a Hole* – tends to be generally more financially successful than other emotional arcs. Furthermore, this relative success is apparent irrespective of the movie genre and does not dependent on the movie production budget. If we assume that IMDb rating can be used as a proxy of viewer satisfaction, we can also conclude that the *Man in a Hole* emotional arc tends to succeed not because it generates movies which are most desired by the public (i.e., achieve the highest ratings on IMDb), but because movies with this emotional arc tend to be most unusual and spark debate. In other words, the *Man in a Hole* emotional arc tends to generate most "talked about" movies and not necessarily "most liked" movies and thereby achieve higher revenues than movies in other categories.

What are the implications of our result on the business models of the entertainment industry? On the one hand, it may appear that when evaluating movie scripts, motion picture production companies should opt for scripts offering *Man in a Hole* emotional journeys. Yet, on the other hand, this would be an oversimplification of our results. We show that when emotional arcs are combined with different genres and produced in different budget categories any of the 6 emotional arcs may produce financially successful films. Therefore, a careful selection of the *script-budget-genre* combination will lead to financial success. It is obvious, however, that data science can significantly advance the dialog between motion picture production companies and the viewers and help generate "on demand", customer-centric, and even personalized content which consumers of motion pictures would be interested in purchasing. The sentiment analysis of movies as an essential part of the business model choice process may shift decision making about desirable content from producers to consumers, empowering the viewers to significantly influence (or even shape) motion picture production.

Hollywood is often called the Factory of Dreams. This paper shows that, in its essence, Hollywood is a Factory of Emotions yet, with the help of data science, it may become the Factory of Viewers' Dreams.

Vonnegut, K. (1981) Palm Sunday, Rosetta Books LLC, New York.

# Appendix

**Figure A Screenshot of the *Open Subtitles* Website**

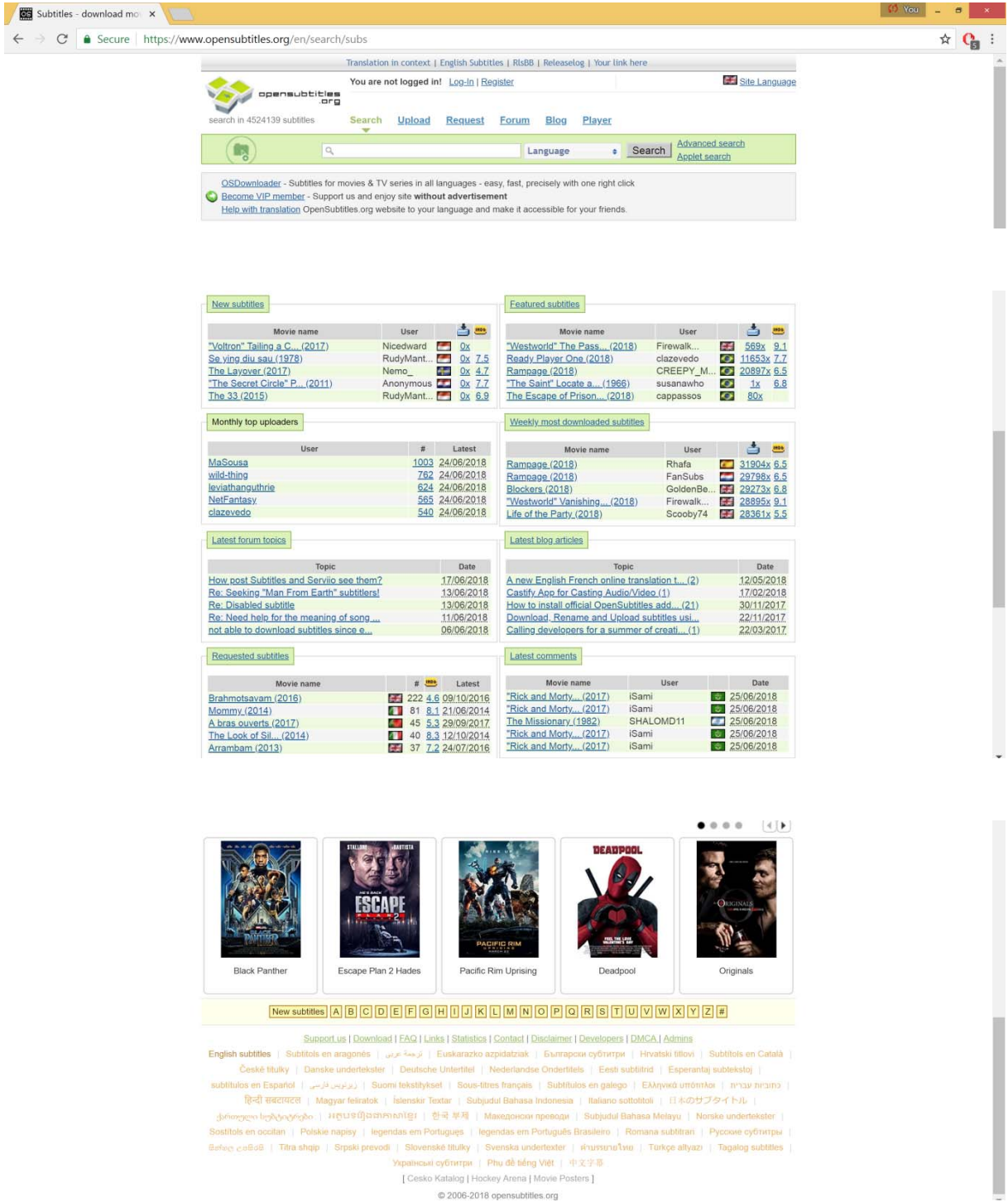

Figure B Screenshot of *The Numbers* Website

**Figure C Screenshot of the *IMDb* Website**

**Table A Results of OLS Regressions for Different Budget Categories and Emotional Arcs**

| Budget (millions) | Arc type | | | | | |
|---|---|---|---|---|---|---|
| | Rags to Riches | Riches to Rags | Man in a Hole | Icarus | Cinderella | Oedipus |
| [0,1]<br>N=107 | -5.224109<br>(5.629633) | -6.142632<br>(4.73962) | 5.027173<br>(4.549392) | .3862553†<br>(5.033921) | 8.620575<br>(5.947797) | -2.800844<br>(6.736746) |
| ]1; 5]<br>N=346 | -2.910223<br>(4.682148) | 1.842595<br>(3.48694) | .2222489<br>(3.336014) | 1.278563<br>(3.522867) | -7.544106†<br>(4.273988) | 5.871116<br>(4.67407) |
| ]5; 10]<br>N=339 | -3.224356<br>(5.295469) | 1.196451<br>(3.528227) | -3.748548<br>(3.47957) | 8.378917*<br>(3.890699) | -3.703672<br>(4.17093) | -.1713286<br>(5.220669) |
| ]10; 20]<br>N=615 | .2205349<br>(5.053811) | -1.301663<br>(3.610417) | -.7713826<br>(3.408389) | 2.825699<br>(3.968776) | 3.160962<br>(4.621801) | -4.742038<br>(5.285587) |
| ]20; 30]<br>N=399 | 5.676744<br>(7.422783) | -7.637175<br>(5.029134) | -.5769682<br>(4.511784) | 1.582348<br>(5.51517) | -2.146044<br>(6.061741) | 12.15966†<br>(7.213049) |
| ]30; 50]<br>N=512 | .5618489<br>(6.837384) | -2.31149<br>(5.446247) | 6.968713<br>(4.49776) | 6.762836<br>(5.77634) | -7.989023<br>(6.140114) | -12.73384†<br>(6.871012) |
| ]50;100]<br>N=518 | -3.242787<br>(9.682033) | -.7803577<br>(7.676915) | 4.329175<br>(6.527757) | -15.59939*<br>(7.941222) | 13.59465<br>(8.997648) | 2.511585<br>(10.09158) |
| ]100;∞[<br>N=215 | -19.20044<br>(29.17517) | 47.83648*<br>(20.84437) | -5.636544<br>(17.75234) | -7.875244<br>(22.71214) | -11.46948<br>(22.98333) | -25.67953<br>(29.15177) |
| All data with budgets<br>(gross domestic revenue)<br>N=3,051 | -2.814695<br>(4.085964) | -1.574803<br>(3.018226) | 5.217438*<br>(2.713389) | -2.643316<br>(3.229983) | .6124211<br>(3.610191) | -2.786345<br>(4.172162) |
| All data with budgets<br>(worldwide revenue)<br>N=3,051 | -9.771201<br>(10.60486) | -3.660632<br>(7.834169) | 12.02102†<br>(7.043771) | -8.366715<br>(8.383287) | 7.670255<br>(9.369617) | -7.494578<br>(10.82919) |
| All data<br>(gross domestic revenue)<br>N=6,174 | -3.233382<br>(2.463638) | -3.459929†<br>(1.782266) | 6.561281***<br>(1.703242) | 2.491377<br>(1.9427) | 1.169003<br>(2.21924) | 1.303107<br>(2.476123) |

Notes: † - significant at 10% level - p<0.1; * - significant at 5% level – p<0.05; ** - significant at 1% level – p<0.01; *** - significant at 0.1% level – p<0.001.

**Table B Results of OLS Regressions for Different Genre Categories and Emotional Arcs**

|  |  | Emotional Arc | | | | | |
|---|---|---|---|---|---|---|---|
|  |  | Rags to Riches | Riches to Rags | Man in a Hole | Icarus | Cinderella | Oedipus |
| IMDb Genre | Action | -425572.4 (1843578) | -656601.6 (1333897) | 2101435 (1275640) | -223427 (1453746) | -341703.3 (1660498) | -1946100 (1852546) |
|  | Horror | -637339 (469180.8) | 748026.9* (339392.2) | 473188.9 (324706.4) | -308567.7 (370004.3) | -478681.3 (422607) | -699972.7 (471489.5) |
|  | SciFi | -2476968† (1502144) | 43956.69 (1087113) | 2627628** (1039304) | -1283523 (1184655) | -194108.2 (1353266) | -797200 (1509878) |
|  | Mystery | -1699425* (811172.9) | 834583.1 (587035.4) | 1127753* (561417.9) | 74742.67 (639872.4) | -1063525 (730751.8) | -1068948 (815364.4) |
|  | Thriller | -2656055* (1271649) | 1999508* (920072.7) | 2584076** (879787.4) | -547898.1 (1003081) | -1403310 (1145630) | -3986405** (1277388) |
|  | Animation | -460890.7 (1093829) | -1462182† (791230.7) | 1637497* (756748.2) | -1539628† (862322.2) | 1256144 (985088.2) | 770403.6 (1099215) |
|  | Drama | 383817 (1403499) | -1377468 (1015354) | 620436.1 (971316.4) | 98940.47 (1106727) | -20935.95 (1264129) | (826833.6 1410415) |
|  | Adventure | -2687334 (2048764) | -2599832† (1482217) | 4361248** (1417033) | -1589129 (1615640) | 2712143 (1845241) | -2261555 (2058991) |
|  | Fantasy | -1932498 (1633923) | -1327524 (1182235) | 3173946** (1130222) | -2709719* (1288105) | 1935933 (1471623) | -188535.6 (1642195) |
|  | Crime | -502808.5 (898884.6) | 514115.9 (650370.1) | 997198.2 (621992.2) | 722297 (708768.6) | -1619651* (809376.8) | -1743385† (903084.3) |
|  | Comedy | -818404.2 (1526535) | -4770596*** (1102877) | 4062499*** (1055253) | -2041299† (1203488) | 3459727* (1374266) | 472224.5 (1534115) |
|  | Romance | 1356745 (1135975) | -1085874 (821911.7) | -148644.6 (786281.7) | -1083663 (895763.9) | 539118.8 (1023260) | 2126881† (1141409) |
|  | Family | -1085045 (1386149) | -1388151 (1002840) | 2943775** (958650.4) | -2484058* (1092636) | 2137875† (1248259) | -1060047 (1393014) |
|  | Biography | 2004688*** (481708.7) | -362366.2 (348997.7) | -279703.1 (333832.5) | -415974.1 (380344.3) | 214510.4 (434471.2) | -327221.8 (484753.9) |
|  | Sport | -584826 (398657.6) | 71963.78 (288496.9) | -261261.9 (275933.6) | -98971.5 (314411.4) | -172642.2 (359123.9) | 1378340*** (400316.5) |
|  | Music | 810496.3* (357268.5) | -552740* (258513.1) | -345437.7 (247324.7) | -64397.97 (281838.2) | 433717.8 (321875.9) | 540752.2 (359120.6) |
|  | War | 796213.2† (457354.1) | 94754.25 (330996.8) | -76360.46 (316604.8) | 118273.6 (360729.3) | -685708.7† (411944.1) | -164358.9 (459726) |
|  | Western | -46941.95 (255716.7) | 324409.3† (184977.8) | -16847.32 (176978.4) | -200519.9 (201628.3) | -302613.8 (230290.6) | 159520.8 (256975.2) |
|  | History | 927585.9* (391581.2) | -541673† (283372.8) | 10194.04 (271131.2) | -30459.71 (308919.2) | -231489.1 (352842.3) | 424307.3 (393661.3) |
|  | Musical | -148088.3 (568563.5) | 455990.1 (411344.5) | 466015.7 (393452.6) | -230174.3 (448330.3) | -744302.9 (512016.1) | -412240 (571356.9) |
|  | Film Noir | -1261.251 (2463.749) | 601.0122 (1782.663) | -893.1598 (1705.126) | 3003.127 (1942.44) | -1301.649 (2219.066) | -1259.66 (2475.945) |
|  | News | n.a. | n.a. | n.a. | n.a. | n.a. | n.a. |

Notes: † - significant at 10% level - $p<0.1$; * - significant at 5% level – $p<0.05$; ** - significant at 1% level – $p<0.01$; *** - significant at 0.1% level – $p<0.001$.